\documentclass[final]{cvpr}

\usepackage{times}
\usepackage{epsfig}
\usepackage{graphicx}
\usepackage{amsmath}
\usepackage{amssymb}
\usepackage{booktabs}
\usepackage{multirow}
\def\confYear{CVPR 2022}
\usepackage[pagebackref=true,breaklinks=true,colorlinks,bookmarks=false]{hyperref}
\usepackage{multirow}

\begin{document}

\title{Perceiving and Modeling Density is All You Need for Image Dehazing}

\author{
    Tian Ye$^{1*}$,
    Mingchao Jiang$^{2*}$,
    Yunchen Zhang$^{3}$ \thanks{Equal contribution.$^\dag$Corresponding author.},
    Liang Chen$^{4}$,
    Erkang Chen$^{1\dag}$,
    Pen Chen$^1$,
    Zhiyong Lu$^2$
    \\
    $^1$School of Ocean Information Engineering,
    Jimei University, Xiamen, China\\
    $^2$JOYY AI GROUP,
    Guangzhou, China\\
    $^3$China Design Group Co., Ltd.
    Nanjing, China\\
    $^4$Fujian Provincial Key Laboratory of Photonics Technology, Fujian Normal University, Fuzhou, China. \\

}

\maketitle
\begin{abstract}
In the real world, the degradation of images taken under haze can be quite complex, where the spatial distribution of haze is varied from image to image. Recent methods adopt deep neural networks to recover clean scenes from hazy images directly. However, due to the paradox caused by the variation of real captured haze and the fixed degradation parameters of the current networks, the generalization 
ability of recent dehazing methods on real-world hazy images is not ideal. 
To address the problem of modeling real-world haze degradation, we propose to solve this problem by perceiving and modeling density for uneven haze distribution.

We propose a novel Separable Hybrid Attention (SHA) module to encode haze density by capturing features in the orthogonal directions to achieve this goal. 
Moreover, a density map is proposed to model the uneven distribution of the haze explicitly. The density map generates positional encoding in a semi-supervised way—such a haze density perceiving and modeling capture the unevenly distributed degeneration at the feature level effectively. Through a suitable combination of SHA and density map, we design a novel dehazing network architecture, which achieves a good complexity-performance trade-off.

The extensive experiments on two large-scale datasets demonstrate that our method surpasses all state-of-the-art approaches by a large margin both quantitatively and qualitatively, boosting the best published PSNR metric from 28.53 dB to 33.49 dB on the Haze4k test dataset and from 37.17 dB to 38.41 dB on the SOTS indoor test dataset.
\end{abstract}

\section{Introduction}
Single image dehazing aims to generate a haze-free image from a hazy image. It is a classical image processing problem, which has been an important research topic in the vision communities within the last decade \cite{ffa-net,wu2021contrastive,2021ntire,ancuti2020ntire}. Numerous real-world vision tasks (e.g., object detection and auto drive) require high-quality clean images, and the fog and haze usually lead to degraded images. Therefore, it is of great interest to develop an effective algorithm to recover haze-free images.

\begin{figure}
    \centering
    \includegraphics[height=3.7cm]{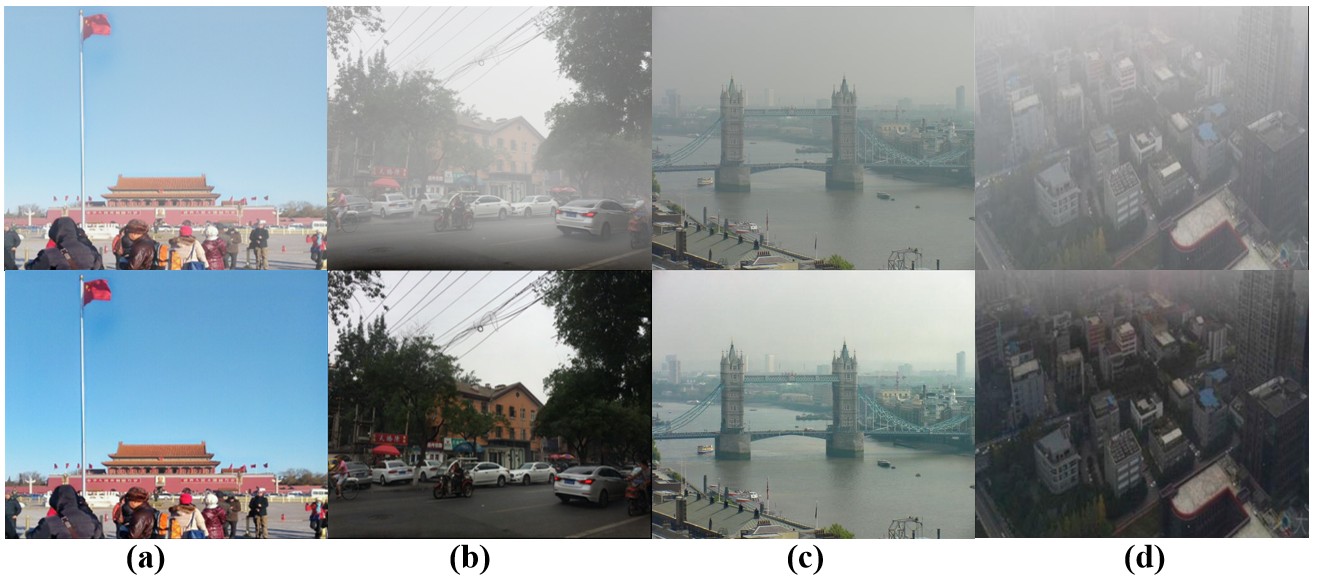}
    \caption{Examples of our image dehazing. Hazy images are in the top row, of which (a) and (b) are from synthetic hazy dataset haze4k \cite{liu2021synthetic}, (c) and (d) are Real-world hazy images from web. The corresponding dehazed results are in the bottom row.}
    \label{fig:banner_examples}
\end{figure}

Haze is a common atmospheric phenomenon in our daily life. Images with haze and fog lose details and color fidelity. Mathematically, the image degradation caused by haze could be formulated by the following model:
\begin{equation}
   \mathbf{I}(\mathbf{x})=\mathbf{J}(\mathbf{x}) t(\mathbf{x})+\mathbf{A}(1-t(\mathbf{x}))
\label{eq:hazy model}
\end{equation}
where $\mathbf{I}(\mathbf{x})$ is the hazy image, $\mathbf{J}(\mathbf{x})$ is the clear image, 
$t(\mathbf{x})$ is the transmission map, $\mathbf{A}$ is the global atmospheric light.

Methods \cite{he2010single, fattal2014dehazing,berman2016non,jiang2017image} based on priors first estimate the above two unknown parameters $t(\mathbf{x})$ and $\mathbf{A}$, then inversely solve the above infinitive formula.
Unfortunately, these handcrafted priors do not always hold in diverse real-world hazy scenes, resulting in inaccurate estimated $t(\mathbf{x})$. 

\begin{figure*}[t]
    \centering
    \includegraphics[width=15cm]{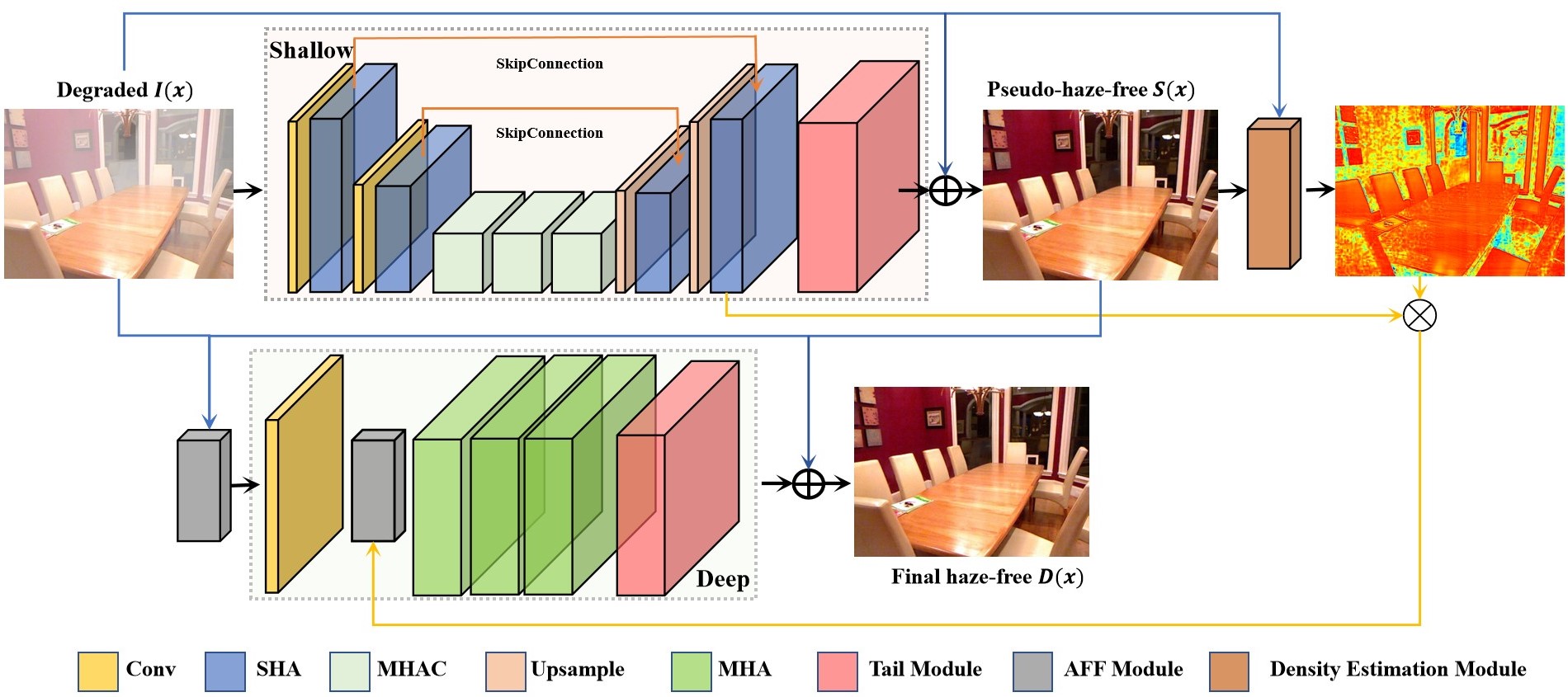}
    \caption{The Overview of network architecture.
The shallow layers is used to make the density map joined with the density estimation module, 
which consists of a stack of MHAC blocks and a Tail Module.
The Deep layers emphasize on detailed reconstruction by Adaptive Features Fusion (AFF) module and Multi-branch Hybrid Attention Block (MHA).
We give details of the structure and configurations in Section 3.}
    \label{fig:SHA-Net}
\end{figure*}

Therefore, in recent years, more and more researchers have begun to pay attention to data-driven learning algorithms \cite{aod,msbdn,kddn,wu2021contrastive,ffa-net}. Different from traditional methods, deep learning methods achieve superior performance via training on large-scale datasets. However, deep learning based algorithms still has the following challenges: (1) \emph{Number of parameters: huge.} The previous methods \cite{liu2020trident,msbdn,yu2021two} boost their performance by increasing the capacity of models which result in a large number of parameters. 
Lack of  consideration of the characteristics of haze also leads to the application failure when encountering practical scenes. (2) \emph{Training Strategy: Complex.} Complicated loss functions \cite{kddn,wu2021contrastive,yu2021two,wu2020knowledge,liu2021synthetic} and fancy training strategies are design to better optimize the reconstruction process. However, these designs lead to high training costs and poor convergence. (3) \emph{Speed: slow.} Some recent methods \cite{ffa-net} extract the feature maps of hazy images under full resolution in every model stage in order to achieve effective performance, thus slowing the speed in the practical test.

 To address the above problems, we formulate the problem in terms of a perceiving and modeling density for uneven haze distribution. From the Eq.\ref{eq:hazy model}, we can notice that the haze degradation model is highly associated with absolute position of image pixels. The key to solve the dehazing problem lies in correctly encoding the haze intensity with its absolute position. Therefore, we propose a network to model the density of haze distribution in an end-to-end manner, which encodes the co-relationship between haze intensity and its absolute position. And our method has a powerful ability to restore image details and color fidelity on real domain and synthetic domain images, as seen in Fig. \ref{fig:banner_examples}.



We propose the method from three different levels: primary block of the network, architecture of the network and map of haze density information to refine features: 

$\bullet$ \textit{Primary Block:} We propose an efficient attention mechanism to perceive the uneven distribution of degradation of features among channel and spatial dimensions: Separable Hybrid Attention (SHA), which effectively samples the input features through a combination of different pooling operations, and strengthens the interaction of different dimensional information by channel scaling and channel shuffle. Our SHA based on horizontal and vertical encoding can obtain sufficient spatial clue from input features.

$\bullet$ \textit{Density Map:} Density Map is a coefficient matrix that encodes the co-relationship between haze intensity and absolute position. The density map explicitly models the intensity of the haze degradation model at corresponding spatial locations. The density map is obtained in an end-to-end matter, semantic information of the scene is also introduced implicitly, which makes the density map more consistent with the actual distribution.

$\bullet$ \textit{Network Architecture:} We design a novel network architecture that restores hazy images with the coarse-to-fine strategy. The architecture of our network mainly consists of three parts: shallow layers, deep layers and density map. We build the Shallow Layers and Deep Layers of our method based on SHA. The shallow layers will generate the coarse haze-free image, which we call the pseudo-haze-free image.
For modeling the uneven degradation of hazy image to refine features explicitly, we utilize the pseudo-haze-free image and the input hazy sample to generate the Density Map. 

Our main contributions are summarized as follows:

\begin{itemize}
    \item We propose SHA as a universal attention mechanism perceives efficiently the degenerated density,
    which can further improve the performance of various image restoration networks.
    \item We propose density map as a novel method to build relationship between different parts in the network, which enhances the coupling of our model. In addition, it can be applied in other low-level vision tasks in the future.
    \item We propose a novel dehazing method based on perceiving and modeling the haze density by highly efficient SHA and density map. Our method achieves the best performance compared with the state-of-the approaches.
\end{itemize}
\section{Related Works}
\subsection{Single Image Dehazing}
Single image dehazing is mainly divided into two categories: a prior-based defogging method \cite{he2010single,colorattenuationprior} and a data-driven method based on deep learning. With the introduction of large hazy datasets \cite{SOTS,liu2021synthetic}, image dehazing based on the deep neural network has developed rapidly. MSBDN \cite{msbdn} uses the classic Encoder-Decoder architecture, but repeated up-sampling and down-sampling operations result in texture information loss. The number of parameters of MSBDN \cite{msbdn} is large, and the model is complex. FFA-Net \cite{ffa-net} proposes an FA block based on channel attention and pixel attention, obtains the final haze-free image by fusing features of different levels. With the help of two different-scale attention mechanisms, FFA-Net has obtained impressive PSNR and SSIM, but the entire model is convolutional operation at the resolution of the original image, resulting in a large amount of calculation and slow speed. It is unfriendly to restore large-size hazy images on devices with low memory. AECR-Net \cite{wu2021contrastive} reuses the FA block and proposes to use a novel loss function based on the contrast learning to make full use of hazy samples, and use a deformable convolution block to improve the expression ability of the model, pushing the index on SOTS \cite{SOTS} to a new high, but memory consumption of the loss function is so high. Compared with AECR-Net \cite{wu2021contrastive}, our model only needs to utilize a simple Charboonier\cite{charbonnier1994two} loss function to achieve higher PSNR and SSIM.
\subsection{Attention Mechanism}
Attention mechanisms have been proven essential in various computer vision tasks, such as image classification \cite{hu2018squeeze}, segmentation\cite{fu2019dual}, dehazing \cite{ffa-net,wu2021contrastive,kddn} and deblurring \cite{zamir2021multi}. One of the classical attention mechanisms is SENet\cite{hu2018squeeze}, which is widely used as a comparative baseline of the plugin of the backbone network. CBAM \cite{woo2018cbam} introduces spatial information encoding via convolutions with large-size kernels, which sequentially infers attention maps along the channel and spatial dimension. Later attention mechanism extends the idea of CBAM \cite{woo2018cbam} by adopting different dimension attention mechanisms to design the advanced attention module. This shows that the key of the performance of attention mechanism is to sample the original feature map fully. There is no effective information exchange between the different dimension attention encoding in before works, thus limiting the lifting of networks. In response to the above problems, we utilize two types of pooling operations to sample the original feature map and cleverly inserted the channel shuffle block in our attention module. Experiments
demonstrate that the performance of our attention mechanism has been dramatically improved due to sufficient feature sampling and efficient feature exchange among different dimensions.
\section{Method}
We propose a novel architecture as shown in Fig.\ref{fig:SHA-Net}, which consists of three parts: shallow layers, deep layers and density map. Shallow layers are responsible for reconstruct high-level contextual content, and deep layers are responsible for rebuilding pixel-level detailed features. 
\subsection{Separable Hybrid Attention Module}
\begin{figure}
    \centering
    \includegraphics[width=8cm]{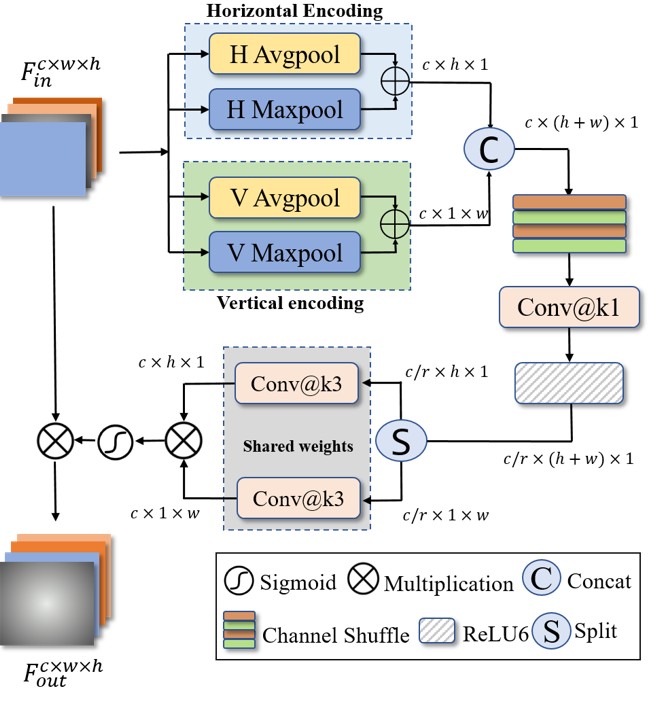}
    \caption{An illustration of the Separable Hybird Attention Module (SHA). Our SHA focuses on the directional embedding, which consists of unilateral directional pooling and convolution. The \textbf{@k} denotes the convolutuion kernel size.}
    \label{fig:sha_block}
\end{figure}
Attention mechanisms based on pooling operations have been applied in many visual tasks and have improved indicators, but previous attention mechanisms such as CBAM \cite{woo2018cbam}, and DANet \cite{fu2019dual} often calculate the attention weights subsequently in the spatial and channel dimensions. But there is no useful information exchanging between the attention of different dimensions. Non-local \cite{wang2018non} and Trilinear Attention \cite{zheng2019looking} have interactive feature-capability but the calculation is very large. So we propose a novel attention mechanism called Separable Hybrid Attention (SHA), which can effectively generate the fine-grained attention weights to guidance the degenerated density perception.
Specifically, the combination of average pooling and maximum pooling not only smoothens the noise in features but also effectively enhances high-frequency information of features. So we utilize the average pooling ($\mathbf{AvgPool}$) and maximum pooling ($\mathbf{MaxPool}$) calculate in the feature $F_{in}$ along the horizontal and vertical directions to obtain the encoding features $v$. The formulas are as follows:
\begin{equation}
\label{eq:avgpool}
v_{avg}^h = \mathbf{AvgPool_h}(F_{in}),
v_{avg}^v = \mathbf{AvgPool_v}(F_{in})
\end{equation}
\begin{equation}
\label{eq:maxpool}
v_{max}^h = \mathbf{MaxPool_h}(F_{in}),v_{max}^v = \mathbf{MaxPool_v}(F_{in})
\end{equation}
We get the direction encode features $v^{h}$ and $v^{v}$  from  Eq.\ref{eq:avgpool} add Eq.\ref{eq:maxpool} correspondingly, so we can make the distribution of features like a normal distribution, which can better reinforce the important information of input features:
\begin{equation}
    v^h = v_{avg}^h + v_{max}^h,v^v = v_{avg}^v + v_{max}^v
\end{equation}
We concatenate the encoded features and utilize the channel shuffle to interaction channel information, aiming to exchange the encoding information between different channels. Then utilizing the 1x1 convolution to reduce the dimension of features, which pass through the nonlinear activation function, so that the features of different channels are fully interactive. The formulas are as follows:
\begin{equation}
\label{eq:interaction}
   [y^h_{c/r},y^v_{c/r}] = \delta(\mathbf{Conv}(\mathbf{cat}([v^h_{c},v^v_{c}])))
\end{equation}
Wherein Eq.\ref{eq:interaction}, the $\delta$ is \textit{ReLU6} activation function, $c$ is the dimensions number and $r$ is the channel scaling factor, $r$ is usually 4.
We use the shared 3x3 convolution to restore the number of channel dimensions for encoded features of different directions, making the isotropic features get similar attention weight values. The final weights can be determined by the larger receptive field of the input features.
\begin{equation}
   y^h_{c} = \mathbf{Conv}(y^h_{c/r}),y^v_{c} = \mathbf{Conv}(y^v_{c/r})
\end{equation}
\begin{figure}[b]
    \centering
    \includegraphics[width=9cm]{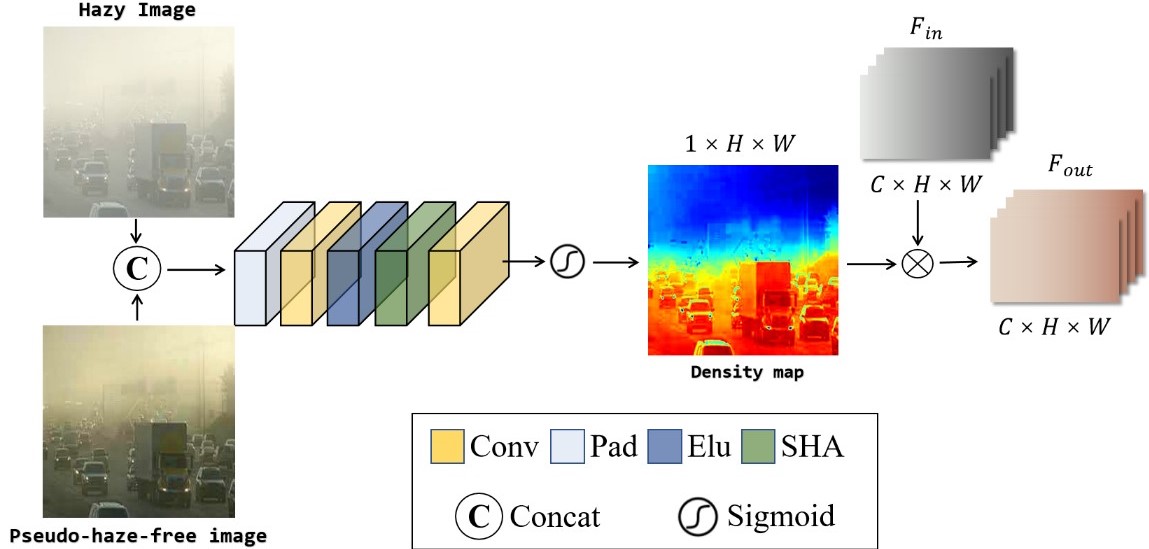}
    \caption{An illustration of the pipeline that generates a density map.}
    \label{fig:supervision_mask}
\end{figure}
Our experiments demonstrate that the 3x3 convolution will help generate a more accurate attention weight matrix. After restoring the channel of the feature, multiply the $y_c^h$ and $y_c^w$ to obtain the attention weight matrix with the same size as the input feature. Finally, a $Sigmoid$ function works to get the attention map $W_{c\times h\times w}$:
\begin{equation}
    W_{c\times h\times w} = \mathbf{Sigmoid}(y^h_{c} \times y^v_{c})
\end{equation}
We multiply the attention weight matrix $W_{c\times h \times w}$ with the input feature $F_{in}$ to get the output feature $F_{out}$:
\begin{equation}
    F_{out} =  W_{c\times h\times w} \otimes F_{in}
\end{equation}
The entire SHA module calculates the attention weight matrix and the process of applying attention weight to the input features, the overview as shown in Fig.\ref{fig:sha_block}. Compared with the previous attention mechanism for image dehazing, our method is less computationally and more effective.

\begin{figure}[t]
    \centering
    \includegraphics[width=8.5cm]{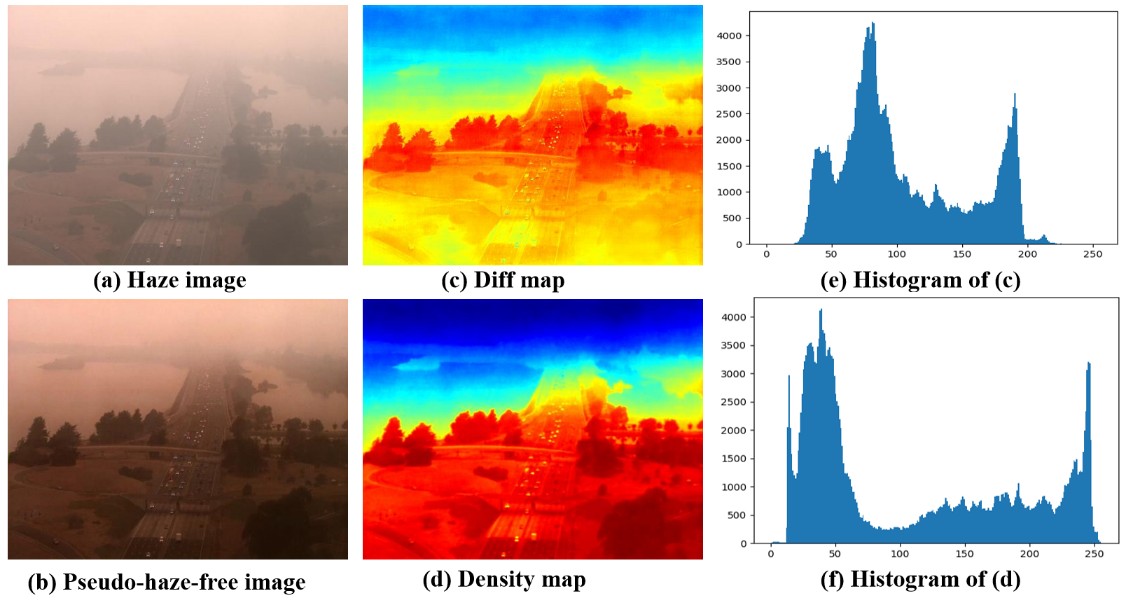}
    \caption{An illustration of the histogram of the diff map and density map. The diff map is the numerical difference between the pseudo-haze-free image and haze images. The value of the density map is mapped from [0,1] to [0,255]. The diff map and density map are visualized by ColorJet for observation and comparison. Note that the histograms of the diff map and density map have similar intensity distributions.}
    \label{fig:histogram}
\end{figure}


\begin{figure*}
    \centering
    \includegraphics[width=11cm]{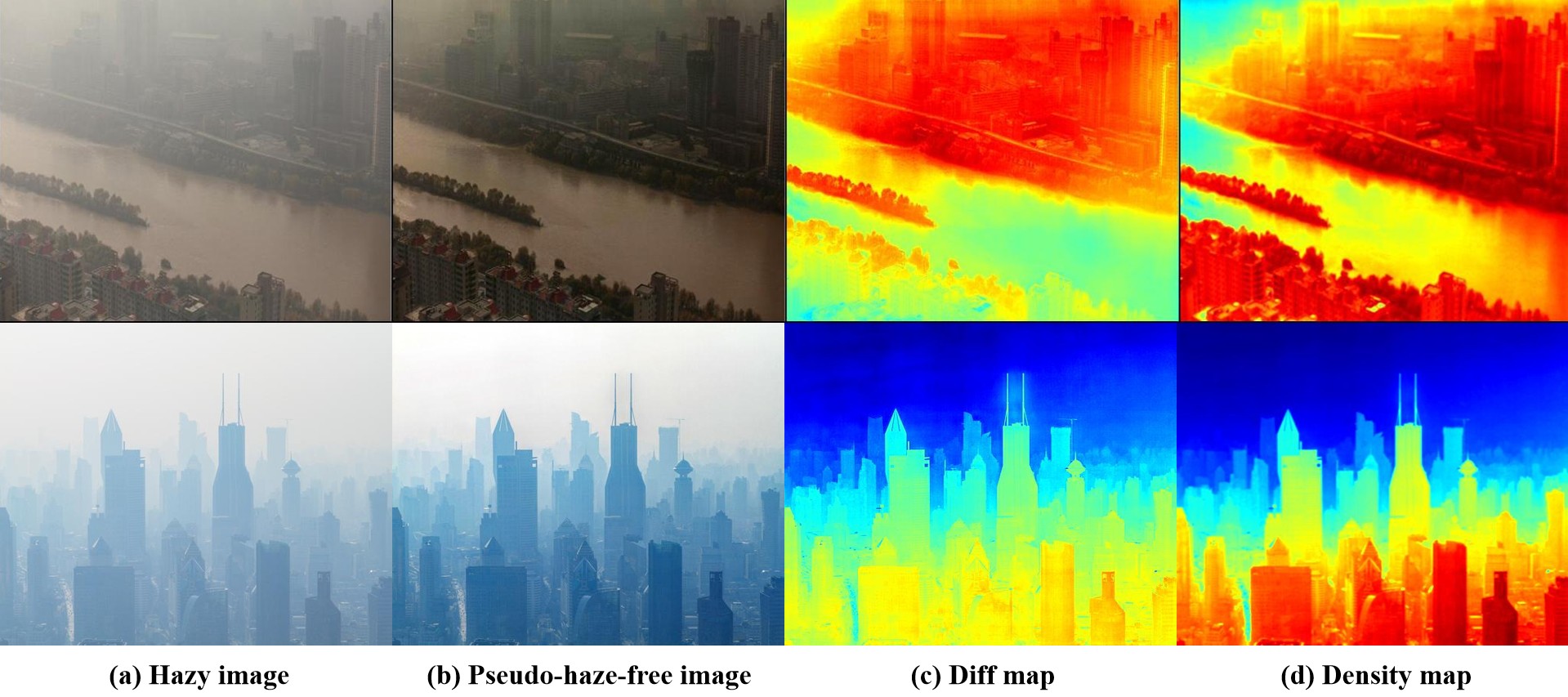}
    \caption{Visual comparisons on real-world hazy images, the diff map is the difference between the pseudo-haze-free and hazy images. It's worth noting that our density map effectively models the density of the degradation spatially compared with the diff map.}
    \label{fig:densitymap_comparison}
\end{figure*}
\subsection{Shallow Layers}
\begin{figure}[b]
    \centering
    \includegraphics[width=8cm]{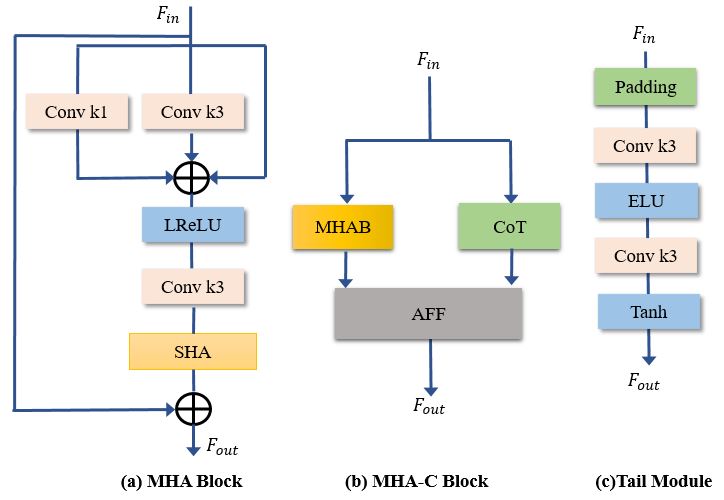}
    \caption{An illustration of the basic modules of the network. (a) The Multi-branch Hybrid Attention Block (MHAB). (b) The MHAC block of shallow layers. (c) The Tail Module.}
    \label{fig:BasicBlock}
\end{figure}

The shallow layers is stacked by several Multi-branch Hybrid Attention with COT Block and the Encoding-Decoding context module. We use the shallow layers to generate the pseudo-haze-free image, which has high-level semantic information.
\paragraph{Multi-branch Hybrid Attention.}
We design the Multi-branch Hybrid Attention (MHA) Block, mainly consisting of the SHA module with parallel convolution. The multi-branch design will improve the expressive ability of the network by introducing multi-scale receptive-filed. The multi-branch block comprises parallel 3x3 convolution, 1x1 convolution, and a residual connection, as shown in Fig.\ref{fig:BasicBlock} (a). 
The degradation is often uneven in space in degraded images, such as hazy images and rainy images, and the spatial structure and color of some areas in the picture are not affected by the degradation of the scene, so we set the local residual learning to let the feature pass the current block directly without any process, which also avoids the disappearance of the gradient.
\paragraph{Adaptive Features Fusion Module.}
We hope that the network can adjust the proportion of feature fusion adaptively according to the importance of different information. Different from the MixUP \cite{wu2021contrastive} module in AECR-Net, we utilize the Adaptive Features Fusion Module to combine two different blocks. The formula is as follows:
\begin{equation}
\begin{split}
    F_{out} &=\mathbf{AFF}(\mathbf{block1},\mathbf{block2}) \\
            &= \sigma (\theta) * \mathbf{block1} + \sigma (1-\theta) *\mathbf{block2}     \\
\end{split}
\end{equation}
Wherein, $block$ denotes the block module, which has the same output size, $\sigma$ is the $Sigmoid$ activation function, and $\theta$ is a learnable factor.

\paragraph{Multi-branch Hybrid Attention with CoT.}
Long-range dependence is essential for feature representation, so we introduce an improved CoT \cite{li2021contextual} block combined with the MHA block in the shallow layers to mine the long-distance dependence of sample features that further expand the receptive field.
Specifically, we design a parallel mix block that uses the MHA block and the improved CoT block to capture local features and global dependencies simultaneously. To fusion the attention result, an Adaptive Features Fusion module is followed back as shown in Fig.\ref{fig:BasicBlock} (b), we call this MHAC block. The formulas are as follows:
\begin{equation}
\begin{split}
    F_{out} &=\mathbf{AFF}(\mathbf{MHAB}(F_{in}),\mathbf{CoT}(F_{in})) \\
            &= \sigma (\theta) * (\mathbf{MHAB}(F_{in}) + \sigma (1-\theta) *\mathbf{CoT}(F_{in})     \\
\end{split}
\end{equation}
The $F_{in}$ denotes the input features, $F_{out}$ is the adaptive mixing results from MHA and CoT block.
Considering that the BN layer will destroy the internal features of the sample, we use the IN layer to replace the BN layer in the improved CoT Block and use the ELU as the activation function. The basic block of shallow layers.

\paragraph{Tail Module.}
As shown in Fig.\ref{fig:BasicBlock}(c), we design the Tail module, which fusions the extracted features and restores the hazy image. The $tanh$ activation function is often used on degradation reconstructions. So we use that as the activation of the output after a stack of 3x3 convolution.

\paragraph{Architecture of Shallow Layers.}
As shown in Fig.\ref{fig:SHA-Net}, we use the shallow layers to reconstruct the degraded image context content. In order to effectively reduce the amount of calculation and expand the receptive field of the convolution of MHAC, we utilize 2 convolutions with a stride of 2 to reduce the resolution of the feature map to 1/4 of the original input firstly; each convolution follows a SHA module. Then use a stack of 8 MHAC blocks with 256 channels, which have 2 skip connections to introduce shallow features before up-sampling, and utilize the Tail module for obtaining the residual of the restored image of the shallow layers:
\begin{equation}
    S(x) = \mathbf{Shallowlayers}(x) + x
\end{equation}
Wherein the $S(x)$ denotes the pseudo-haze-free image, the $x$ denotes the hazy input image.
\subsection{Density Map}

To alleviate the problem of uneven haze distribution in the spatial position, we adopt the density map to perceive the implicit spatial correlation between pseudo-haze-free images generated by shallow layers and hazy images. 

Like all classical deep learning methods, our approach adopts convolutions to encode input to high-level feature representations that map the position information to the activation area onto high-level feature maps. Unlike the classical deep learning methods that only utilize information from the hazy image, the proposed density map is based on fully exploring the correlation between the pseudo-haze-free image and the hazy image. 

As is depicted in Fig. \ref{fig:histogram}, it is obvious that shallow convolution network has the ability to reconstruct a pseudo haze-free image. But it's weak to model complex spatial distribution that is disturbed by contextual information variation, which leads to an inaccurate estimation of the degradation model.

Since hand-crafted prior may not model spatial co-relationship between image pairs effectively, we adopt a simple convolution network to estimate a density map that shares the same spatial dimensions as input. As the illustration in Fig. \ref{fig:supervision_mask}, Density Estimation Module splices the pseudo-haze-free image and the hazy sample of the shallow layers in the channel dimension. We expand the features to 64 channels using the 3x3 convolution after utilizing the Reflected Padding to avoid the detail loss of the feature edge. And then utilize the SHA module to explore perceiving the uneven degeneration of input features fully, finally using a convolution operation to compress the shape of the feature. The sigmoid function is used to get the density map $M \in \mathbb{R}^{1 \times H \times W}$.After getting the density map $M$, we multiply $M$ by the input feature $F_{in}\in \mathbb{R}^{C \times H \times W}$ to get the final output $F_{out}\in \mathbb{R}^{C \times H \times W}$:

\begin{equation}
 F_{out} = F_{in} \otimes M   
\end{equation}

As is depicted in Fig. \ref{fig:histogram} and Fig. \ref{fig:densitymap_comparison}, the visualization of density map clearly demonstrate that our density map effectively model the uneven haze distribution spatially. The diff map is the numerical difference between the pseudo-haze-free and haze input images, which can be considered a degenerated layer of the hazy veil. The intensity of the density map changes with the depth of scenes and hazy distribution.
\subsection{Deep layers}
\label{sec:deep_layers}
A lot of degraded image restoration methods \cite{DMPHN,zamir2021multi,msbdn} often use the fully Encoder-Decoder structure to learn a large receptive fields. However, repeated down-sampling and up-sampling operations can easily cause the loss of texture details, which will greatly affect the visual perception of the image and the natural effect, so we utilize the deep layers to repair the pixel-level texture details with the supervised signal from density map. As shown in Fig. \ref{fig:SHA-Net}, we utilize 10 MHA blocks with 16 channels to extract features at the resolution of the original input. In order to avoid unnecessary calculations caused by repeated extraction of features, we use the AFF module to introduce and mix features refined by our density map from the shallow layers adaptively.

\section{Experiments}

\subsection{Datasets and Metrics}
We chose the PSNR and SSIM as experimental metrics to measure the performance of our SHA-Net.We train on two large synthetic datasets, RESIDE \cite{SOTS} and Haze4k \cite{liu2021synthetic}, and testing on SOTS \cite{SOTS} and Haze4k \cite{liu2021synthetic} testing sets, respectively. The indoor training set of RESIDE \cite{SOTS} contains 1,399 clean image and 13,990 hazy images generated by corresponding clean images. The indoor testing set of SOTS contains 500 indoor images. The training set of Haze4k \cite{liu2021synthetic} contains 3,000 hazy images with ground truth images and the testing set of Haze4k \cite{liu2021synthetic} contains 1,000 hazy images with ground truth images.

\subsection{Training Settings}
We augment the training dataset with randomly rotated by 90,180,270 degrees and horizontal flip. The training image patches with the size $256 \times 256$ are extracted as input $I_{in}$ of our network.
The network is trained for $7.5 \times 10^5$, $1.5 \times 10^6$ steps on Haze4k \cite{liu2021synthetic} and RESIDE \cite{SOTS} respectively.We use Adam optimizer with initial learning rate of $2 \times 10^{-4}$, and adopt the CyclicLR to adjust the learning rate, where on the triangular mode, the value of gamma is 1.0,base momentum is 0.8, max momentum is 0.9, base learning rate is initial learning rate and max learning rate is $3 \times 10^{-4}$. PyTorch \cite{paszke2017automatic} was used to implement our models with 4 RTX 3080 GPU with total batchsize of 40.
\subsection{Loss Function}
We only use Charbonnier loss \cite{charbonnier1994two} as our optimization objective:
\begin{equation}
\mathcal{L}(\Theta)= \mathcal{L}_{\text {char }}(S(x),J_{gt}(x))+ \mathcal{L}_{\text {char }}(D(x),J_{gt}(x)))
\end{equation}
Where $\Theta$ denotes the parameters of our network, the $S(x)$ denotes the pseudo-haze-free image, $D(x)$ denotes the output of deep layers, which is the final output image, $J_{gt}$ stands for ground truth, and $\mathcal{L}_{\text {char }}$ is the Charbonnier loss \cite{charbonnier1994two}:
\begin{equation}
\mathcal{L}_{\text {char }}=\frac{1}{N} \sum_{i=1}^{N} \sqrt{\left\|X^{i}-Y^{i}\right\|^{2}+\epsilon^{2}}
\end{equation}
with constant $\epsilon$ emiprically set to $1e^-3$ for all experiments.
\subsection{Ablation Study}

\begin{table}[h]
	\centering
	\caption{Comparisons on Haze4k testset for different configurations of Shallow Layers.}
	\label{tab:ablation_on_shallowlayers}
		\begin{tabular}{cccccc}
			\toprule
			Model &&&&& \cr \hline
			Base &\checkmark &\checkmark &\checkmark &\checkmark &\checkmark \cr 
			+FA  & &\checkmark & & & \cr
			+SHA & & &\checkmark &\checkmark &\checkmark\cr
			+CoT & & & &\checkmark &\checkmark\cr
			+AFF & & & & &\checkmark \cr
			\midrule
			PSNR &24.40 & 25.30 &26.39 &26.84 &27.28 \cr
			\bottomrule
		\end{tabular}
\end{table}
To demonstrate the effectiveness of the proposed method, we conduct ablation study to analyze different elements, including SHA, CoT, AFF and different configurations of our network.
We first construct our \textit{base} Network as the baseline of shallow layers of dehazing network, which lack SHA, CoT, and AFF. Subsequently, we add the different modules into the baseline as:
(1) \textbf{base+SHA:} Add the Separable Hybrid Attention module into baseline.
(2) \textbf{base+FA:} Add the Feature Attention \cite{ffa-net} module into baseline.
(3) \textbf{base+SHA+CoT:} Add both SHA and CoT operation into baseline.
(4) \textbf{base+SHA+CoT+AFF:} The complete combination of our shallow layers.

We employ the Charbonnier loss \cite{charbonnier1994two} as training loss function for ablation study, and utilize Haze4k \cite{liu2021synthetic} dataset for both training and testing. The performance of above models are summarized in Table \ref{tab:ablation_on_shallowlayers}.

~\\
 \textbf{Effect of Separable Hybrid Attention.}
Separable Hybrid Attention module significantly improves the performance from Base to Base+SHA with an increase of 1.99 dB PSNR. Therefore, SHA is a significant component due to the high-performance gain. We also evaluate the performance of the FA \cite{ffa-net} module with the base model as show in Table \ref{tab:ablation_on_shallowlayers}.

We compared the Flops and Parameters of SHA module with other universal and dehazing attention modules, such as the Feature Attention module \cite{ffa-net} used in dehazing networks \cite{ffa-net,wu2021contrastive,wu2020knowledge}, and the common CBAM \cite{woo2018cbam},as shown in Table \ref{tab:attention_comparison}.Compared with other attention module used in image dehazing networks, the Flops of our SHA module is lower.

The result demonstrates that our attention mechanism has better performance because of the excellent feature coding mechanism.

\begin{table}[h]
	\centering
	\caption{The Flops and Params comparison of universal attention mechanism and dehazing network. The SWRCA is the attention module of KDDN\cite{kddn}.}
	\label{tab:attention_comparison}
		\begin{tabular}{cccccc}
			\toprule
			Attention &Flops(M) &Params \\
			\midrule
			SE\cite{hu2018squeeze} &4.195 &512 \\
			ECA\cite{wang2020eca} &4.195 &3 \\
			CBA\cite{woo2018cbam} &10.619 &1.122K \\
			\midrule
			FA\cite{ffa-net} &38.864 &1.625K \\
			SWRCA\cite{kddn} &2424.311 &41.088K \\
			SHA(Ours) &15.29 &5.192K \\
			\bottomrule
		\end{tabular}
\end{table}
\begin{figure*}[]
     \centering
     \includegraphics[width=17cm]{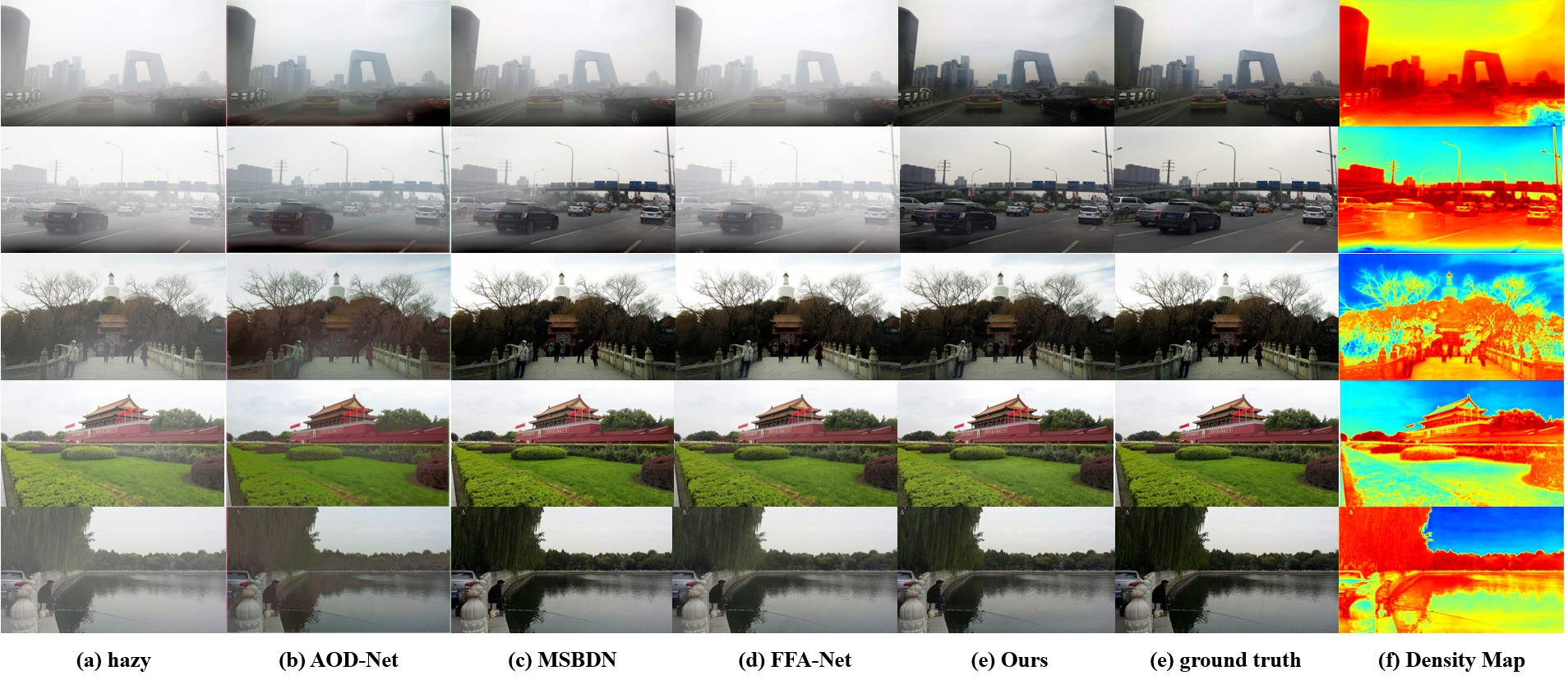}
     \caption{Visual comparisons on dehazed results of various methods on synthetic hazy images and our density map, the images from haze4k \cite{liu2021synthetic} test dataset. The images are best viewed in the full-screen mode.}
     \label{fig:synthetic comparison}
\end{figure*}

~\\
\textbf{Study on Separable Hybrid Attention.}
We verify the effect of main operations of Separable Hybrid Attention we proposed. As shown in Table \ref{tab:ablation_sha_components}, we demonstrate that the combination of $\textbf{AvgPool}$ and $\textbf{MaxPool}$ is nontrivial for the performance of attention mechanism we proposed. The Channel Shuffle operation boost the encoding information exchanging among different channels, which is essential for an efficient and effective attention mechanism.
\begin{table}[h]
	\centering
	\caption{Comparisons on Haze4k testset for different configurations of Separable Hybrid Attention.}
	\label{tab:ablation_sha_components}
		\begin{tabular}{cccccc}
			\toprule
			AvgPool&\checkmark &\checkmark&\checkmark \cr
			MaxPool& &\checkmark&\checkmark \cr
			Channel Shuffle& & &\checkmark& \cr
			\midrule
			PSNR &28.28 &31.16 &33.49 \cr
			\midrule
			Conv@k1 &\checkmark && \cr
			Conv@k3 & &\checkmark& \cr
		    \midrule
		    PSNR &32.21 &33.49 & - \cr
		    
			\bottomrule
		\end{tabular}
\end{table}

We also verify the suitable kernel size of the last convolution of SHA, as shown in Table \ref{tab:ablation_sha_components}; we choose the 3x3 convolution as the final processing operation for parameter-performance trade-off.

\begin{table}[h]
	\centering
	\caption{Comparisons on Haze4k testset for different configurations of our methods.}
	\label{tab:ablation2}
		\begin{tabular}{cccccc}
			\toprule
			Shallow Layers&\checkmark &\checkmark&\checkmark \cr
		    Deep Layers& &\checkmark&\checkmark \cr
			density map& & &\checkmark& \cr
			\midrule
			PSNR &27.28 &30.33 &33.49 \cr
			\bottomrule
		\end{tabular}
\end{table}
\textbf{Effect of Deep Layers and Density Map.}
We verify the effect of deep layers of our method. As described in section \ref{sec:deep_layers}, the deep layers effectively relieve the texture detail loss by repeated sampling. The deep layers significantly improve the model performance with an increase of 3.05 dB PSNR, as shown in Table \ref{tab:ablation2}. The result demonstrates that the feature extracted processing with the original image resolution is essential for pixel-level texture detail restoration. And we also verify the effect of density map we proposed, as shown in Table \ref{tab:ablation2} and  Fig.\ref{fig:densitymap_comparison}. It's worth noting that our density map effectively predicts the hazy density of hazy input, which models the uneven distribution of hazy degradation and improves the performance with an increase of 3.16 dB PSNR.




\section{Compare with SOTA Methods}
\paragraph{Visual Comparison.}
To validate the superiority of our method, as shown in Fig.\ref{fig:synthetic comparison}, firstly we compare the visual results of our method with previous SOTA methods on synthetic hazy images from Haze4k \cite{liu2021synthetic} dataset. It can be seen that the other methods are not able to remove the haze in all the cases, while the proposed method produced results close to the real clean scenes visually. Our method is superior in the recovery performance of image details and color fidelity. Please refer to the supplementary materials for more visual comparisons on the synthetic hazy images and real-world hazy images.
\begin{table}
		\centering
			\caption{Quantitative comparisons of our models with the state-of-the-art dehazing methods on Haze4k \cite{liu2021synthetic} and SOTS \cite{SOTS} datasets (PSNR(dB)/SSIM). Best results are \underline{underlined}.}
			\label{tab:comparisons}
			\resizebox{8cm}{!}{
		\begin{tabular}{l|c|c|c|c}
		    \toprule[1.2 pt]
			\multirow{2}{*}{Method}             &      \multicolumn{2}{c|}{Haze4k \cite{liu2021synthetic}}       &       \multicolumn{2}{c}{SOTS \cite{SOTS}}        \\[1pt]
			                                    & {PSNR$\uparrow$}  & {SSIM$\uparrow$}  & {PSNR$\uparrow$}  & {SSIM$\uparrow$}  \\[1pt] \hline
		
			DCP \cite{he2010single}  &       14.01       &       0.76       &         15.09         &         0.76         \\[1pt]
			
			NLD\cite{berman2016non}     &15.27 &0.67 &17.27 &0.75 \\[1pt]
			
			DehazeNet \cite{cai2016dehazenet}     &       19.12       &       0.84       &       20.64       &       0.80       \\[1pt]
			AOD-Net \cite{aod}                    &       17.15       &       0.83       &       19.82       &       0.82       \\[1pt]
			GDN \cite{griddehazenet}     &         23.29         &         0.93         &       32.16       &       0.98       \\[1pt]
			MSBDN \cite{msbdn}       &       22.99       &       0.85       &       33.79       &       0.98       \\[1pt]
			FFA-Net \cite{ffa-net}                  &       26.96       &       0.95       &       36.39       &       0.98       \\[1pt]
			AECR-Net \cite{wu2021contrastive}                  &       -       &       -       &  37.17   &      {0.99}      \\[1pt]
			DMT-Net \cite{liu2021synthetic}                   &       28.53       &       0.96       &       -       &       -       \\[1pt] \hline
			\textbf{Ours}   & \underline{33.49} & \underline{0.98} &       \underline{38.41}       & 0.99 \\[1pt] 
			\bottomrule[1.2 pt]
		\end{tabular}
		}
\end{table}
\paragraph{Quantitative Comparison.}
We compare quantitatively the dehazing results of our SHA-Net with SOTA single image dehazing methods on Haze4k \cite{liu2021synthetic} and SOTS \cite{SOTS} datasets. As shown in Table \ref{tab:comparisons}, the SHA-Net outperforms all SOTA methods, achieving 33.49dB PSNR and 0.98 SSIM on Haze4k \cite{liu2021synthetic}. It increases the PSNR by 4.93dB compared to the second best method. On SOTS \cite{SOTS} indoor test set, the SHA-Net also outperforms all SOTA methods, achieving 38.41dB PSNR and 0.99 SSIM. It increases the PSNR by 1.24dB, compared to the second best method.


\section{Conclusion}
In this paper, we propose a powerful image dehazing method to recover haze-free images directly. Specifically, the Separable Hybrid Attention is design to better perceive haze density and a density map is to further refine extracted features. Although our method is simple, it is superior to all the previous state-of-art methods with a very large margin on two large-scale hazy datasets. Our method has a powerful advantage in the restoration of image detail and color fidelity. We hope to further promote our method to other low-level vision tasks such as deraining, super-resolution, denosing and desnowing.

\textit{Limitations:} The proposed method recovers high-fidelity haze-free images. As is shown in Fig.\ref{fig:synthetic comparison}, haze free images are usually in low-light mode, which is not visually pleasant in real-life scenes. Following the main idea of this work, future research can be made in various aspects to generate high-quality images with pleasant visual perceptions.

{\small
\bibliographystyle{ieee_fullname}
\bibliography{egbib}
}

\end{document}